\documentclass[twocolumn]{article}

\usepackage[top=25mm,bottom=30mm,left=20mm,right=20mm]{geometry}
\usepackage{titlesec}
\usepackage{graphicx}
\usepackage{caption}
\usepackage{footnote}
\usepackage{url}
\usepackage{indentfirst}
\usepackage{times}
\usepackage{multicol}
\usepackage{amsmath}
\usepackage{amsfonts}
\usepackage{amssymb}
\usepackage{float}
\usepackage{multirow}
\usepackage{bm}

\usepackage{color}
\newcommand{\red}[1]{\textcolor{red}{#1}}
\newcommand{\blue}[1]{\textcolor{blue}{#1}}

\titleformat{\section}{\fontsize{12pt}{12pt}\bfseries}{\thesection}{1em}{}
\titleformat{\subsection}{\fontsize{10.5pt}{10.5pt}\bfseries}{\thesubsection}{1em}{}

\makeatletter
\def\affiliation#1{\def\@affiliation{#1}}
\renewcommand{\maketitle}{
  \begin{center}
    {\fontsize{12pt}{12pt}\bfseries \@title\par}
    \vskip 1ex
    {\fontsize{10.5pt}{10.5pt} \@author\par}
    \vskip 1ex
    {\fontsize{10.5pt}{10.5pt} \@affiliation\par}
  \end{center}
}
\makeatother

\title{Efficient Cost-and-Quality Controllable Arbitrary-scale Super-resolution \\ with Fourier Constraints}
\author{Kazutoshi Akita* and Norimichi Ukita**}
\affiliation{*Toyota Technological Institute, sd21501@toyota-ti.ac.jp \\ **Toyota Technological Institute, ukita@toyota-ti.ac.jp}

\begin{document}

\twocolumn[\begin{@twocolumnfalse}
\maketitle

\vspace{1em}

\begin{center}
\begin{minipage}{0.8\textwidth}
\noindent
\textbf{Abstract:}  
Cost-and-Quality (CQ) controllability in arbitrary-scale super-resolution is crucial. Existing methods predict Fourier components one by one using a recurrent neural network. However, this approach leads to performance degradation and inefficiency due to independent prediction. This paper proposes predicting multiple components jointly to improve both quality and efficiency.

\end{minipage}
\end{center}

\vspace{1em}

\end{@twocolumnfalse}]

\section{Introduction}
\label{sec:intro}

Super-resolution (SR), which reconstructs high-resolution (HR) images from low-resolution (LR) ones, has become an essential technique in intelligent systems. In manufacturing, SR can enhance visual inspection by revealing subtle surface defects that may otherwise be overlooked~\cite{DBLP:journals/tim/KondoU24}. In autonomous driving of vehicles and drones, SR enables more reliable detection of small or distant objects captured by the onboard cameras~\cite{DBLP:conf/iconip/HarisSU21,DBLP:conf/mva/FujiiAU21,DBLP:journals/access/AkitaU23}, thereby improving safety. These applications demonstrate that SR serves as a key enabling technology in mechatronic perception systems, since accurate recognition of fine details must be achieved.

While deep learning-based SR methods~\cite{srcnn,vdsr,rcan,DBLP:conf/cvpr/TimofteAG0ZLSKN17,ntire2018,dbpn,DBLP:journals/pami/HarisSU21} have achieved remarkable success, most conventional approaches are designed for fixed integer scale factors (e.g., x2, x4). Such fixed-scale SR methods are insufficient for some real-world mechatronic systems, where the required scale factor often varies depending on the application scenario or sensor configuration. To overcome this limitation, arbitrary-scale SR (ASSR) methods have recently been developed, enabling SR at any desired scale factor in one SR model, including non-integer factors. Representative previous methods, like LIIF~\cite{liif} and LTE~\cite{lte}, achieve arbitrary scaling by formulating SR as a continuous function defined on image coordinates. In this formulation, the SR image can be reconstructed by querying pixel values at arbitrary image coordinates, allowing the one SR model to produce SR images at any resolution. Although these ASSR methods have expanded the applicability of SR, they focus on improving the reconstruction quality of SR images. In real-world scenarios, however, improving quality alone is not sufficient. Many practical systems often face diverse computational conditions, where the computational budgets for SR may vary depending on both accuracy and available computational resources. This situation calls for a new property beyond scale flexibility: Cost-and-Quality (CQ) controllability.

\begin{figure}[t]
    \centering
    \includegraphics[width=\linewidth]{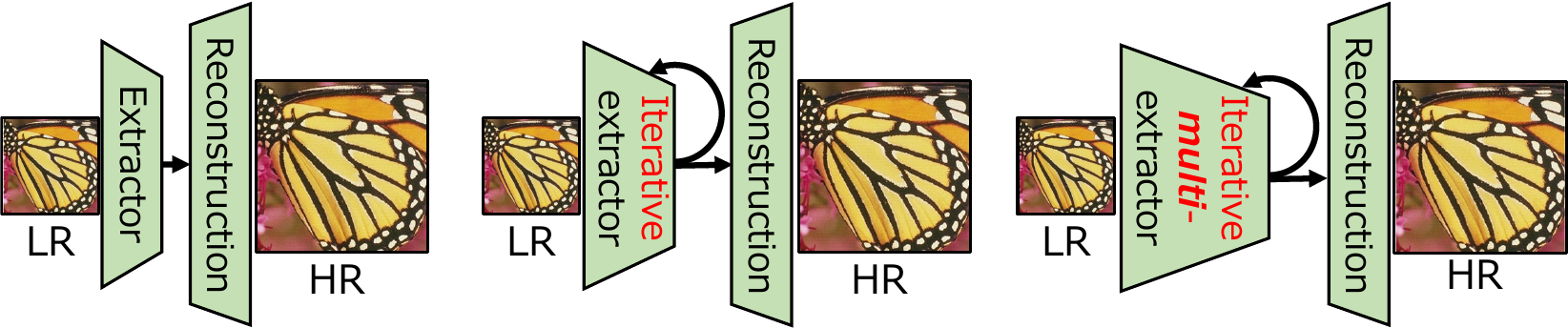}
    \begin{minipage}{0.24\linewidth}
    {\small (a) Non CQ-controllable}
    \end{minipage}
    \hspace*{1mm}
    \begin{minipage}{0.34\linewidth}
    {\small (b) CQ control with single-feature predictions~\cite{our}}
    \end{minipage}
    \hspace*{1mm}
    \begin{minipage}{0.34\linewidth}
    {\small (c) CQ control with multi-feature predictions (Ours)}
    \end{minipage}
    \vspace*{2mm}
    \caption{
       Comparison of SR frameworks. (a) Non-CQ-controllable methods reconstruct HR images with a fixed cost–quality trade-off in one trained model. (b) RecurrentLTE achieves CQ controllability by predicting Fourier components one by one through an RNN, but suffers from limited accuracy and inefficiency. (c) The proposed method predicts multiple Fourier components jointly at each recurrence, improving both reconstruction accuracy and efficiency while preserving CQ controllability.
    }
    \label{fig:teaser}
    \vspace*{-4mm}
\end{figure}

CQ controllability refers to the ability of an SR method to control the trade-off between computational cost and reconstruction quality at inference time. By dynamically adjusting this trade-off, the same model can operate in lightweight settings when computational resources are limited or when high-quality reconstruction is not critical, or in high-quality settings when resources are available or when accurate reconstruction is critical. This flexibility is particularly important in applications such as autonomous driving, where different on-board devices may have different camera qualities and computational capabilities. A single CQ-controllable SR model can accommodate these diverse conditions, operating efficiently on resource-limited platforms while also providing high-quality outputs on more powerful devices to ensure reliable recognition of small distant objects. Hence, CQ controllability is an essential property for bringing SR methods closer to practical deployment.

To achieve both arbitrary scaling and CQ controllability, RecurrentLTE~\cite{our} employs a recurrent neural network (RNN) to sequentially predict Fourier components, each consisting of amplitude and frequency, one by one, as shown in Fig.~\ref{fig:teaser} (b). In this method, each recurrence of the RNN predicts one Fourier component. When more recurrences are used, more components are predicted, and the reconstruction quality is improved. Conversely, using fewer recurrences reduces the number of predicted components, leading to more efficient inference. In this way, users can control the trade-off between computational cost and SR quality with a single model. 

However, RecurrentLTE underperforms compared to non-CQ-controllable methods even with a sufficient number of recurrences. This performance degradation may be attributed to one-by-one prediction, where each Fourier component is predicted independently. Since Fourier components collectively represent image details, ignoring their correlations limits the reconstruction accuracy. In addition, the sequential prediction process is inefficient, as the runtime grows linearly with the number of recurrences required for high-quality reconstruction. These limitations highlight the need for a more accurate and efficient framework for CQ-controllable ASSR.

To address these limitations, this paper proposes an RNN-based framework that predicts multiple Fourier components jointly at each recurrence, as shown in Fig.~\ref{fig:teaser} (c). By outputting several components at once, the model can explicitly exploit the dependencies among them, leading to a more accurate reconstruction of image details. At the same time, this joint prediction strategy improves efficiency, since high-quality results can be obtained with fewer recurrences compared to one-by-one prediction. In summary, the proposed method preserves the advantages of CQ controllability and arbitrary scaling while significantly reducing the quality gap with non-CQ-controllable methods.

Our contributions are summarized as follows:
\begin{itemize}
    \item Our CQ-controllable arbitrary-scale SR method predicts multiple Fourier components jointly in each recurrence of the RNN.
    This simultaneous prediction reduces the number of recurrences, leading to efficient SR reconstruction.
    \item While multi-component learning is more difficult than single-component learning, our method suppresses this difficulty by explicitly constraining the interdependency between components predicted at each recurrence. This constraint allows our method to maintain the SR image quality comparable to that of single component prediction~\cite{our}.
    \item Compared to the base method~\cite{our}, in our method, inference time reduces in inverse proportion to the number of jointly predicted components while maintaining quality.
\end{itemize}

\section{Related Work}
\label{sec:related_work}

\subsection{Super-resolution}
\label{subsec:sr}

As discussed in Sec.~\ref{sec:intro}, early deep learning-based SR methods~\cite{srcnn,vdsr,rcan,DBLP:conf/cvpr/TimofteAG0ZLSKN17,ntire2018} are restricted to fixed integer scale factors (e.g., $\times 2$, $\times 4$). Such a limitation reduces their applicability in real-world scenarios where arbitrary scales are often required. To overcome this, arbitrary-scale SR (ASSR) methods have been proposed. MetaSR~\cite{metasr} dynamically generates upsampling filters to support any scale factor. SRWarp~\cite{srwarp} combines results from multiple integer-scale SR outputs to approximate arbitrary scales. LIIF~\cite{liif} adopts an implicit image function that predicts pixel values at continuous coordinates, enabling flexible scaling. LTE~\cite{lte} extends LIIF by incorporating Fourier representations, allowing high-frequency details to be reconstructed more effectively. Although these approaches enable arbitrary scaling within a single model, they lack cost-and-quality (CQ) controllability.

To jointly achieve arbitrary scaling and CQ controllability, an RNN-based implicit representation is proposed in RecurrentLTE~\cite{our}. This method sequentially predicts Fourier components to reconstruct an SR image. By adjusting the number of recurrences at inference time, users can control the trade-off between computational cost and reconstruction quality. However, its one-by-one prediction scheme is inefficient and fails to exploit the dependencies among Fourier components, which are important for accurate reconstruction.

In this study, we address this limitation by introducing an RNN-based framework that predicts multiple Fourier components jointly at each recurrence. By modeling component dependencies explicitly, the proposed method improves reconstruction accuracy for high-frequency details such as edges, while also enhancing computational efficiency.

\subsection{CQ-controllable Neural Networks}

CQ controllability in neural networks has been explored through several approaches.  
One representative method is knowledge distillation~\cite{distillation1,distillation2}, where a teacher network transfers its knowledge to a smaller student network that is optimized for specific conditions, such as limited computational resources.  
However, this strategy is impractical for end users, as it requires retraining student networks whenever conditions change.  

Another line of work involves pruning~\cite{pruning1,pruning2} and quantization~\cite{quantize1,quantize2}, which compress a trained model by removing redundant parameters or reducing weight precision.  
Similar to distillation, these methods rely on additional steps such as data-driven pruning or fine-tuning~\cite{quantizeSV,pruningSV}, and thus cannot be applied flexibly at inference time by end users.  

A more direct approach to CQ controllability is the early-exit strategy, which enables test-time adaptation without additional training.  
By inserting intermediate output layers, the inference process can be terminated early depending on available computational resources.  
For example, Huang et al.~\cite{huang} proposed an image classification model with intermediate classifiers in DenseNet, and Larsson et al.~\cite{larsson} introduced a network with a fractal structure that allows flexible exits.  

Building on this idea, RecurrentLTE~\cite{our} achieves CQ-controllable arbitrary-scale SR by adjusting the number of RNN recurrences.
While this framework maintains CQ controllability, its one-by-one Fourier component prediction remains inefficient.  
In contrast, our method predicts multiple Fourier components jointly, thereby improving both reconstruction accuracy and computational efficiency.

\section{CQ-controllable Arbitrary-scale SR}
\label{sec:method}

\begin{figure}[t]
    \centering
    \includegraphics[width=\linewidth]{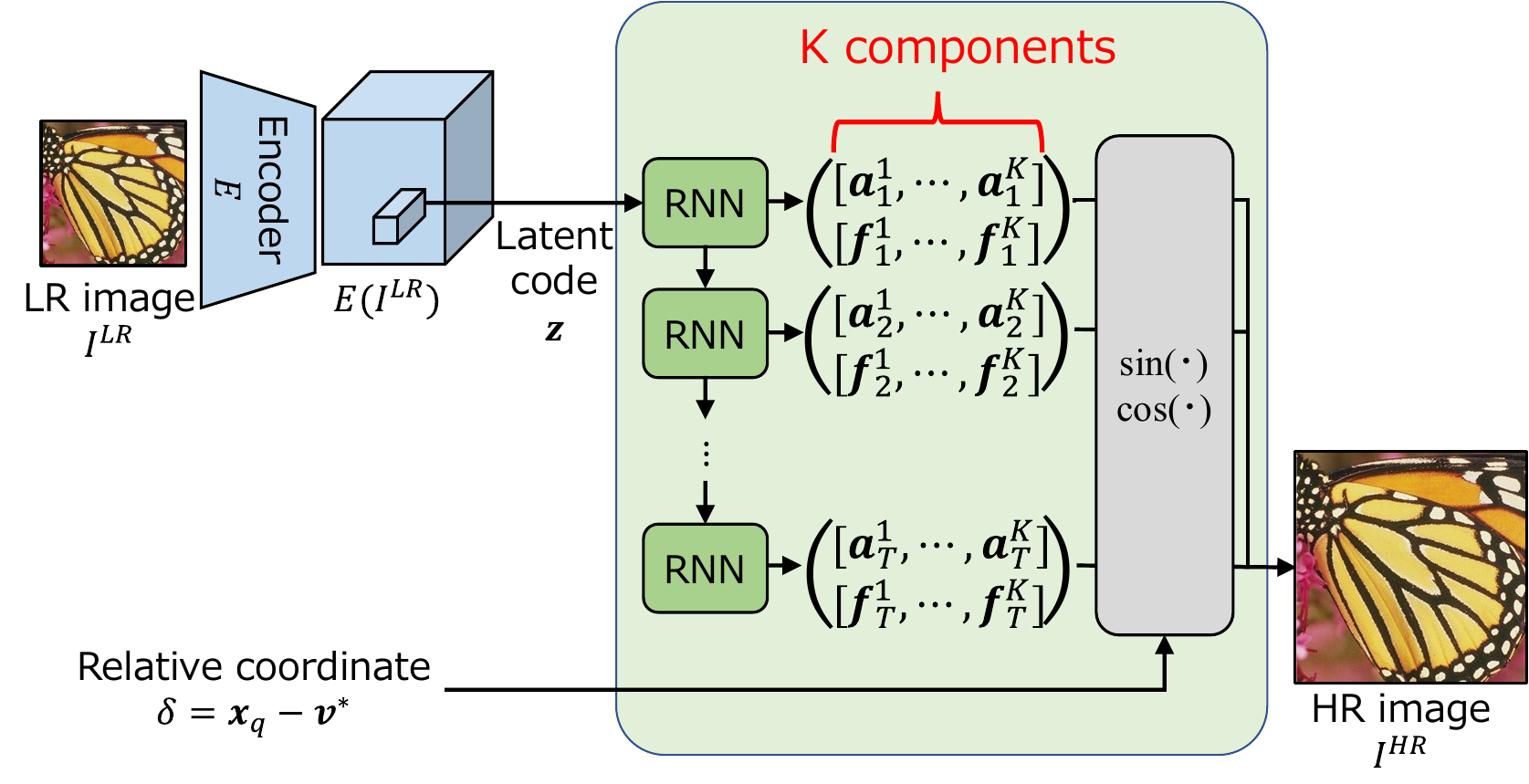}
    \caption{Overview of our CQ-controllable arbitrary-scale SR method. 
    Following LIIF~\cite{liif}, a latent code $z$ is extracted from an input LR image using an encoder $E$.
    The code $z$ is fed into an RNN to predict Fourier components, as in RecurrentLTE~\cite{our}.
    Unlike~\cite{our}, our method predicts multiple $K$ Fourier components at each recurrence.
    These predicted components are then used to reconstruct the HR image.}
    \label{fig:overview}
    \vspace*{-3mm}
\end{figure}

An overview of our method is illustrated in Fig.~\ref{fig:overview}.  
Our approach extends RecurrentLTE~\cite{our}, which itself builds on LIIF-based arbitrary-scale SR methods.  
For clarity, Sec.~\ref{subsec:liif} reviews LIIF-based methods, Sec.~\ref{subsec:our} explains RecurrentLTE, and Sec.~\ref{subsec:propose} introduces our proposed framework.

\subsection{LIIF-based Arbitrary-scale SR}
\label{subsec:liif}

LIIF-based SR methods~\cite{liif,lte,our} reconstruct an SR image $I^{SR}$ by predicting RGB values for each query pixel position $\bm{x}_q$, which is a continuous coordinate in the HR coordinate system.
This prediction is conditioned on a latent code $\mathbf{z}$ extracted from an LR image $I^{LR}$ using an encoder $E$.
Let $\phi$ be the implicit function, then the RGB value at $\bm{x}_q$ is predicted as:
\begin{equation}
    I^{SR}(\bm{x}_q) = \phi(\mathbf{z}, \delta),
    \label{eq:liif}
\end{equation}
where $\delta = \bm{x}_q - \bm{v}^{*}$ is the relative position of the query $\bm{x}_q$ to the coordinate $\bm{v}^{*}$ of the latent code $\mathbf{z}$ in the HR coordinate system. 
By evaluating $\phi$ at arbitrary coordinates, SR images at arbitrary scales can be reconstructed.

To enhance reconstruction accuracy, LTE~\cite{lte} and RecurrentLTE~\cite{our} adopt Fourier representations.  
Specifically, Eq.~(\ref{eq:liif}) is reformulated with sinusoidal functions as:
\begin{equation}
    I^{SR}(\bm{x}_q) = \mathbf{A} \odot
    \begin{bmatrix}
        \cos(\pi \mathbf{F}\delta) \\
        \sin(\pi \mathbf{F}\delta)
    \end{bmatrix},
\label{eq:sum}
\end{equation}
where $\mathbf{A}$ and $\mathbf{F}$ are amplitude and frequency vectors predicted by a Fourier estimator $\psi$:
\begin{equation}
    \mathbf{A}, \mathbf{F} = \psi(\mathbf{z}). 
\end{equation}

\subsection{CQ-controllable RNN for Arbitrary-scale SR}
\label{subsec:our}

To achieve CQ controllability, RecurrentLTE~\cite{our} employs an RNN as $\psi$, which predicts one Fourier component (amplitude $\bm{a}_t$ and frequency $\bm{f}_t$) at each recurrence $t$:
\begin{eqnarray}
    \bm{a}_t, \bm{f}_t, \mathbf{h}_t &=& {\rm RNN}(\bm{a}_{t-1}, \bm{f}_{t-1}, \mathbf{h}_{t-1}), \label{eq:single_fourier} \\
    \mathbf{A} &=& [\bm{a}_1, \cdots, \bm{a}_T]^{\top}, \nonumber \\
    \mathbf{F} &=& [\bm{f}_1, \cdots, \bm{f}_T]^{\top}, \nonumber
\end{eqnarray}
where $\mathbf{h}_t$ is the hidden state and $T$ is the number of Fourier components used for reconstruction.  
At inference time, $T$ can be freely chosen in the range $1 \leq T \leq T_{\rm max}$, 
where $T_{\rm max}$ is a hyperparameter that specifies the maximum number of recurrences used during training.
In this paper, we set $T_{\rm max}$ to 60.
Thus, end users can dynamically adjust the number of recurrences $T$ to control the trade-off between computational cost and reconstruction quality.  

The total network (Fig.~\ref{fig:overview}) is trained with the pixel-wise L1 reconstruction loss over all $N_{p}$ pixels of the SR image:
\begin{equation}
    \mathcal{L}_{i} = \frac{1}{N_{p}} \sum_{p=1}^{N_{p}} \left| I^{SR}_{p} - I^{HR}_{p} \right|.
    \label{eq:image_loss}
\end{equation}

\subsection{Proposed Method: Joint Prediction of Multiple Fourier Components}
\label{subsec:propose}

\begin{figure}[t]
    \centering
    \includegraphics[width=\linewidth]{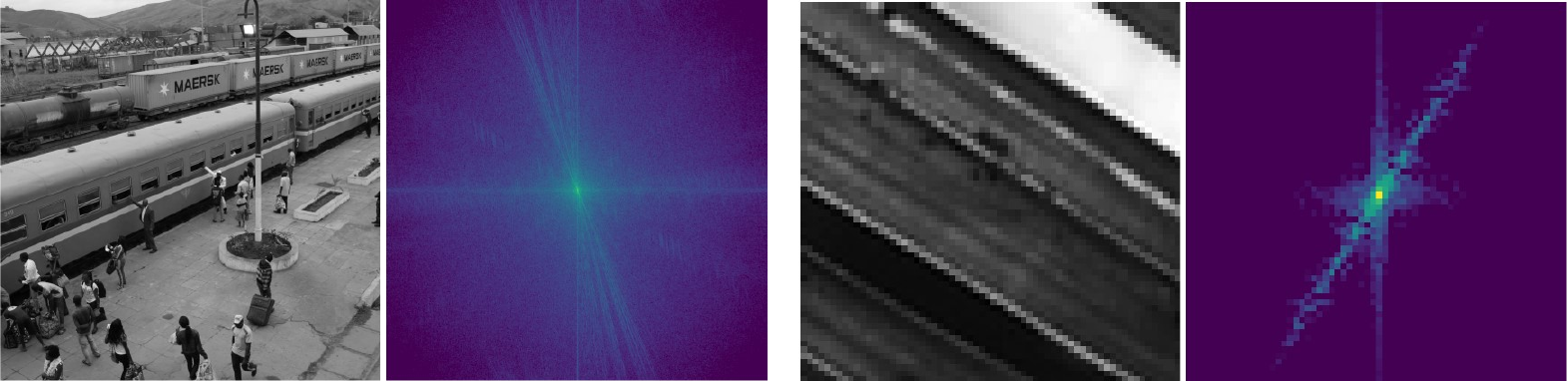}
    (a) Original image \hspace*{9mm} (b) Small patch\\
    \vspace*{0mm}
    \caption{Directional bias in Fourier components. In (a) and (b), the left and right images show an image and its amplitude spectrum, respectively.  
    Edge-biased patterns in the spatial domain correspond to aligned Fourier components along specific directions in the frequency domain.}
    \label{fig:fourier}
    \vspace*{-3mm}
\end{figure}

RecurrentLTE predicts only one Fourier component at each recurrence, so dependencies among components, which are important for reconstructing fine structures such as edges and textures, are not explicitly modeled.  
To address this limitation, our method predicts multiple Fourier components jointly at each recurrence.  
Specifically, the RNN outputs $K$ amplitude–frequency pairs at every step $t$:
\begin{eqnarray}
    \mathbf{A}_t, \mathbf{F}_t, \mathbf{h}_t &=& {\rm RNN}(\mathbf{A}_{t-1}, \mathbf{F}_{t-1}, \mathbf{h}_{t-1}), \label{eq:multi_fourier} \\
    \mathbf{A} &=& [\mathbf{A}_1, \cdots, \mathbf{A}_{\frac{T}{K}}]^{\top}, \nonumber \\
    \mathbf{F} &=& [\mathbf{F}_1, \cdots, \mathbf{F}_{\frac{T}{K}}]^{\top}, \nonumber
\end{eqnarray}
where $\mathbf{A}_t = [\bm{a}_t^1, \cdots, \bm{a}_t^K]^{\top}$ and $\mathbf{F}_t = [\bm{f}_t^1, \cdots, \bm{f}_t^K]^{\top}$.  
For a fixed total number of components $T$, the runtime $C$ is reduced approximately in proportion to $1/K$, since fewer recurrences are required.  

Training multiple components simultaneously is more difficult than predicting them one by one.  
We address this by leveraging a property of Fourier components: when edge structures in the spatial domain are directionally biased, the corresponding Fourier components tend to align along straight lines through the origin in the frequency domain (Fig.~\ref{fig:fourier}).  
This tendency is especially evident in local patches, which are the basic units processed by our model.  
To encourage such alignment among the jointly predicted components, we introduce an additional Fourier alignment loss:
\begin{equation}
    \mathcal{L}_{f} = - \sum_{i,j \in K} \frac{\bm{f}^{i}_{t} \cdot \bm{f}^{j}_{t}}{|\bm{f}^{i}_{t}| \, |\bm{f}^{j}_{t}|}.
    \label{eq:fourier_loss}
\end{equation}

The total training objective is a weighted sum of the pixel-wise reconstruction loss and the Fourier alignment loss:
\begin{equation}
    \mathcal{L} = \mathcal{L}_{i} + w_{f} \mathcal{L}_{f},
\end{equation}
where $w_{f}$ is a balancing weight for Fourier alignment loss.

\section{Experiments}

\subsection{Implementation Details}

The feature extractor $E$ is based on the pretrained backbone of EDSR~\cite{edsr}.  
The RNN is implemented as a 4-layer Linear Transformer~\cite{linear_transformer}, and we use SPE~\cite{spe}, which is a relative position encoding method.
During training, the number of recurrences $T$ is randomly sampled between 1 and $T_{\rm max} (=60)$ for each iteration to ensure CQ controllability, following RecurrentLTE~\cite{our}.  
The scale factors for the horizontal and vertical dimensions ($S_y$ and $S_x$) are independently sampled within the range $[1,4]$. 
Each HR training image is randomly cropped into patches of size $48 S_y \times 48 S_x$, which are downscaled to $48 \times 48$ using bicubic interpolation to generate the corresponding LR patches.  
For each HR patch, 256 query coordinates are randomly sampled.  
The model is trained with a batch size of 16 using the Adam optimizer~\cite{adam} ($\beta_1 = 0.9$, $\beta_2 = 0.999$), with an initial learning rate of $1 \times 10^{-4}$, halved at 200 epochs, for a total of 400 epochs.  
All models are trained on the DIV2K training set~\cite{div2k}.  

Table~\ref{tab:loss_weight} reports the effect of the weight $w_f$ in the training objective $\mathcal{L}_i + w_f \mathcal{L}_f$.  
Based on this result, all subsequent experiments are conducted with $w_f = 10^{-3}$.  

For evaluation, we compare our method with LTE~\cite{lte}, which is non-CQ-controllable, and with RecurrentLTE~\cite{our}, as well as our method with $K=2$ and $K=3$.  
Our method with $K=1$ is identical to RecurrentLTE.  
Unlike our method, LTE predicts a fixed number of Fourier components and is thus inherently non-CQ-controllable.  
To enable a fair comparison, LTE is modified to be CQ-controllable by reconstructing SR images using only the top $T$ Fourier components with the highest amplitudes selected from its fixed set of predicted components.  

All experiments are performed on an NVIDIA Tesla V100 GPU with 32GB of memory.  
We report PSNR (on the luminance channel) as the metric.

\begin{table}[t]
\centering
\caption{PSNR scores of different $w_{f}$ in $\mathcal{L}_{i} + w_{f} \mathcal{L}_{f}$ on the DIV2K validation set. 
The red numbers indicate the \red{best} performance for each scale factor.}
\vspace*{-2mm}
\begin{tabular}{l|cccc} \hline
$w_{f}$ & $10^{-1}$ & $10^{-2}$ & $10^{-3}$ & $10^{-4}$ \\ \hline
$\times 2$ & 34.00 & 34.00 & \red{34.10} & 34.03 \\
$\times 4$ & 28.68 & 28.67 & \red{28.73} & 28.70 \\ \hline
\end{tabular}
\label{tab:loss_weight}
\vspace*{-2mm}
\end{table}

\subsection{Quantitative Results}

Table~\ref{tab:qualitative} shows the PSNR scores of LTE, RecurrentLTE, and our method at scale factors $\times 2$ and $\times 4$.  
The modified LTE exhibits a significant performance drop as the number of Fourier components $T$ decreases.  
In contrast, RecurrentLTE and our method maintain relatively high PSNR scores even with fewer Fourier components.  
Note that in our method, the runtime decreases approximately in inverse proportion to $K$, i.e., about 50\% and 33\% of that of $K=1$ when $K=2$ and $K=3$, respectively.  

Consistent with the findings in~\cite{our}, our methods achieve lower performance than LTE when $T=60$ (i.e., when all Fourier components are used), but outperform LTE at smaller $T$.  
Between $K=1$ and $K=2$, PSNR scores are almost comparable; the average across all $T$ is 33.88 and 33.50 for $K=1$ and $K=2$, respectively.  
These results suggest that predicting multiple components jointly does not suffer from the performance degradation caused by independent one-by-one prediction.  
Conversely, with $K=3$, the performance declines, likely due to overfitting or instability during training.  
This highlights the inherent difficulty of predicting a large number of Fourier components jointly using an RNN, pointing to future work on improved architectures or training strategies.

\begin{table*}[t]
\centering
\caption{PSNR scores for scale factors $\times 2$ and $\times 4$ on the DIV2K validation set.  
For each scale factor and each number of Fourier components $T$, the red and blue values indicate the \red{best} and the \blue{second-best} results among the compared methods, respectively.}

\begin{tabular}{cc||cccccccccc} \hline
\multicolumn{2}{c||}{\multirow{3}{*}{SR methods}} & \multicolumn{10}{c}{\# of Fourier components at test time ($T$)}
\\ \cline{3-12} 
\multicolumn{2}{c||}{}& \multicolumn{5}{c|}{$\times 2$}& \multicolumn{5}{c}{$\times 4$} \\
\multicolumn{2}{c||}{}& 60 & 48 & 36 & 24 & \multicolumn{1}{c|}{12} &  60 & 48 & 36 & 24 & 12 \\ \hline
\multicolumn{2}{c||}{modified LTE~\cite{lte}}                         & \red{34.31} & 22.67 & 18.47 & 22.85 & \multicolumn{1}{c|}{24.15} & \red{28.87} & 22.53 & 18.54 & 22.59 & 23.51  \\
RecurrentLTE~\cite{our}& K=1& 34.06 & \blue{33.98} & \red{34.02} & \red{33.93} & \multicolumn{1}{c|}{\red{33.45}} & 28.63 & \blue{28.57} & \red{28.49} & \red{28.51} & \blue{27.16} \\
\multirow{2}{*}{Ours} & K=2& \blue{34.10} & \red{34.06} & \blue{33.95} & \blue{33.62} & \multicolumn{1}{l|}{\blue{32.16}} & \blue{28.73} & \red{28.62} & \blue{28.54} & \blue{28.38} & \red{27.37}  \\
& K=3& 33.67 & 33.02 & 32.12 & 31.96 & \multicolumn{1}{l|}{30.72} & 27.74 & 27.14 & 26.40 & 25.17 & 24.55  \\ \hline
\end{tabular}
\label{tab:qualitative}
\end{table*}

\subsection{Qualitative Results}

\begin{figure*}[t]
    \centering
    \includegraphics[width=\linewidth]{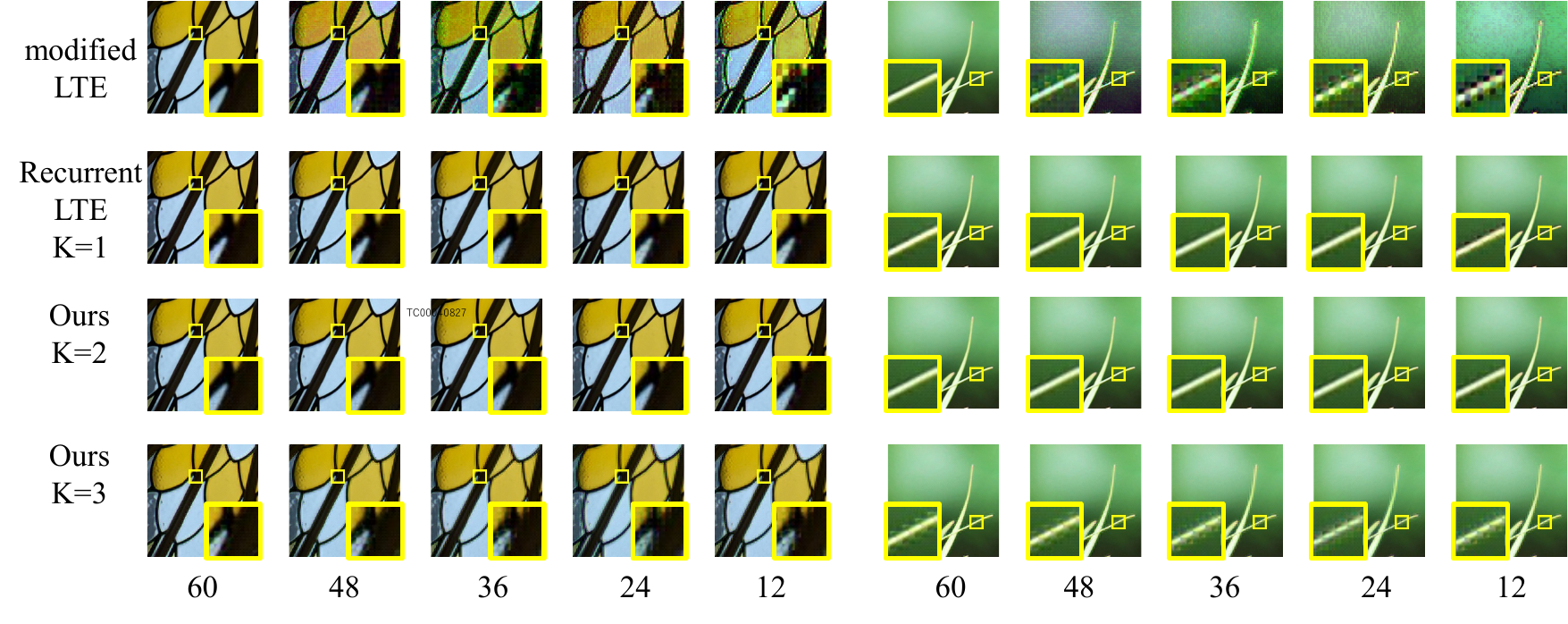}
    \vspace*{-9mm}
    \caption{Qualitative comparison on DIV2K validation set. Rectangular areas are enlarged for better visualization.}
    \label{fig:qualitative}
    \vspace*{-2mm}
\end{figure*}

Qualitative comparisons are presented in Fig.~\ref{fig:qualitative}.  
Our method consistently produces higher-quality images even with fewer Fourier components, whereas LTE exhibits color shifts and artifacts, as highlighted by red boxes (particularly at small $T$).  
These results demonstrate that the proposed architecture generalizes well across different $T$ values and confirms the CQ-controllability of our method.  

Comparing $K=1$ and $K=2$, we observe that $K=2$ produces sharper and more accurate edges, indicating that predicting multiple components jointly is effective for reconstructing fine details.  
However, for $K=3$, the results include unnatural blurs and artifacts, as highlighted by orange boxes.  
This suggests that increasing $K$ may lead to overfitting and reduce the ability to reconstruct fine details accurately.  

\subsection{Detailed Analysis}
\subsubsection{Other Metrics}

To complement the PSNR-based evaluation, we also report results on perceptual quality metrics.  
We adopt the Learned Perceptual Image Patch Similarity (LPIPS)~\cite{lpips}, which measures perceptual similarity to ground-truth HR images (lower is better), and the Naturalness Image Quality Evaluator (NIQE)~\cite{niqe}, a no-reference quality measure that captures perceptual naturalness (lower is better).  
We evaluate these metrics only for $\times 4$ SR, as representative results.  

Table~\ref{tab:perceptual} shows LPIPS/NIQE scores for LTE, RecurrentLTE ($K=1$), and our method with $K=2$ and $K=3$, under different numbers of Fourier components $T \in \{60, 48, 36, 24, 12\}$.  
Consistent with the PSNR results, LTE degrades significantly when $T$ decreases: for example, LPIPS/NIQE increases from 0.299/12.61 at $T=60$ to 0.510/15.37 at $T=12$.  
In contrast, RecurrentLTE maintains stable scores across different $T$, ranging from 0.295/12.47 at $T=60$ to 0.338/13.74 at $T=12$.  
Our method with $K=2$ shows similar robustness, achieving 0.292/12.64 at $T=60$ and still keeping relatively low values (0.372/12.29) even at $T=12$.  
These results confirm that CQ controllability does not compromise perceptual quality.  
Meanwhile, $K=3$ exhibits noticeable degradation, e.g., from 0.325/12.97 at $T=60$ to 0.410/13.24 at $T=12$, indicating that predicting too many components jointly may harm perceptual fidelity.

\begin{table*}[t]
\centering
\caption{Perceptual quality comparison (LPIPS/NIQE, lower is better) at $\times 4$ SR on the DIV2K validation set.}
\begin{tabular}{cc||ccccc} \hline
\multicolumn{2}{c||}{\multirow{2}{*}{SR methods}} & \multicolumn{5}{c}{\# of Fourier components at test time ($T$)} \\ \cline{3-7}
\multicolumn{2}{c||}{} & 60 & 48 & 36 & 24 & 12 \\ \hline
\multicolumn{2}{c||}{modified LTE~\cite{lte}}   & 0.299/12.61 & 0.355/13.33 & 0.372/14.18 & 0.449/13.57 & 0.510/15.37 \\
RecurrentLTE~\cite{our} & K=1                   & 0.295/12.47 & 0.297/13.01 & 0.299/13.17 & 0.311/12.96 & 0.338/13.74 \\
\multirow{2}{*}{Ours}   & K=2                   & 0.292/12.64 & 0.294/12.99 & 0.305/13.26 & 0.316/13.13 & 0.372/12.29 \\
                        & K=3                   & 0.325/12.97 & 0.321/12.79 & 0.348/14.09 & 0.351/14.02 & 0.410/13.24 \\ \hline
\end{tabular}
\label{tab:perceptual}
\end{table*}

\subsubsection{Runtime Analysis}

\begin{figure}[t]
    \centering
    \includegraphics[width=\linewidth]{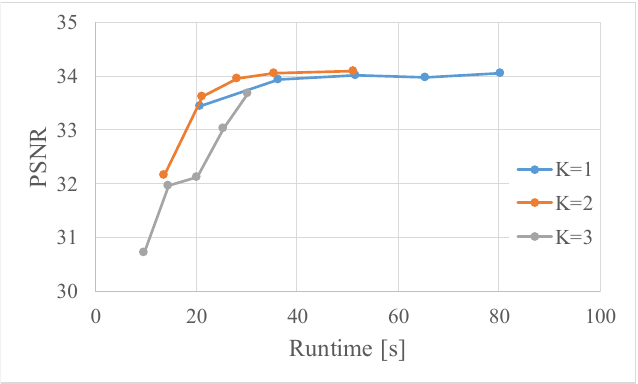}
    \caption{PSNR versus runtime for different numbers of jointly predicted components ($K=1,2,3$). 
    Colors denote $K$. 
    \textbf{Points along each curve correspond to $T=\{60,48,36,24,12\}$ \emph{from right to left}} (i.e., runtime decreases as $T$ decreases). 
    With $K=2$, the curve remains close to $K=1$ while shifting left (approximately half the runtime), whereas $K=3$ further reduces runtime at the expense of accuracy.}
    \label{fig:runtime_psnr}
    \vspace*{-2mm}
\end{figure}

To visualize the effect of CQ controllability on efficiency and accuracy, we plot PSNR against runtime for different values of $K$ in Fig.~\ref{fig:runtime_psnr}. 
For each curve, reducing the number of Fourier components $T$ shifts the point leftwards (lower runtime) and mildly degrades PSNR. 
Compared with RecurrentLTE ($K=1$), our method with $K=2$ achieves almost the same PSNR over a wide range while operating at roughly half the runtime, showing that joint prediction improves efficiency without sacrificing reconstruction quality. 
When $K=3$, the runtime decreases further, but the PSNR drops more noticeably.
Overall, $K=2$ offers the most favorable runtime–quality balance.

\section{Conclusion}

This paper presented a CQ-controllable arbitrary-scale SR method that jointly predicts multiple Fourier components at each recurrence. 
By explicitly modeling dependencies among components, the proposed framework improves the reconstruction of high-frequency details such as edges and textures. 
Extensive experiments showed that our method achieves high-quality SR results while reducing runtime compared to its base method~\cite{our}, thereby confirming that CQ controllability can be preserved without sacrificing efficiency.

Future work includes improving scalability and robustness when predicting multiple components. 
In particular, we observed that performance degrades when $K \geq 3$, even though larger $K$ would in principle allow further runtime reduction by requiring fewer recurrences. 
Addressing this limitation is crucial to unlock the full efficiency potential of joint prediction. 
Moreover, our method underperforms LTE when all Fourier components are used (i.e., $T=60$), suggesting that the current RNN-based formulation is less effective at fully exploiting high-frequency information. 
Future improvements should therefore focus not only on stabilizing joint prediction for larger $K$, but also on enhancing the expressiveness of the predictor at large $T$ so that our method can match or surpass non-CQ-controllable baselines under all settings. 
Another promising direction suggested by our results is to adaptively choose the number of jointly predicted components depending on image content or predicted spectrum, or to incorporate stronger alignment priors. 
Such extensions could stabilize joint prediction beyond $K=2$ and provide more flexible runtime--quality trade-offs.

\bibliographystyle{unsrt}
\bibliography{references}

\end{document}